\documentclass[10pt,twocolumn,letterpaper]{article}

\usepackage{iccv}
\usepackage{times}
\usepackage{epsfig}
\usepackage{graphicx}
\usepackage{amsmath}
\usepackage{amssymb}

\usepackage{subfigure}
\usepackage{color,xcolor}
\usepackage{multirow}
\usepackage{setspace}
\usepackage{url}

\usepackage{bbm}
\usepackage[linesnumbered,ruled]{algorithm2e}

\usepackage[pagebackref=true,breaklinks=true,letterpaper=true,colorlinks,bookmarks=false]{hyperref}

\iccvfinalcopy 

\def\iccvPaperID{8222} 

\ificcvfinal\fi

\begin{document}

\title{Consistency-Aware Graph Network for Human Interaction Understanding}

\title{LAGNet: Logic-Aware Graph Network for Human Interaction Understanding}

\author{Zhenhua Wang$^{~\dagger}$, Jiajun Meng$^{~\dagger}$,  Dongyan Guo$^{~\dagger}$, Jianhua Zhang$^{~\ddagger}$,\\
	Javen Qinfeng Shi$^{~\flat}$, Shengyong Chen$^{~\ddagger}$\\
	$^{\dagger}$~Zhejiang University of Technology; $^{\ddagger}$~Tianjin University of Technology;	$^{\flat}$~The University of Adelaide\\
	{\tt\small zhhwang@zjut.edu.cn}
}

\maketitle 
\ificcvfinal\thispagestyle{empty}\fi

\begin{abstract}
Compared with the progress made on human activity classification, much less success has been achieved on human interaction understanding (HIU). Apart from the latter task is much more challenging, the main cause is that recent approaches learn human interactive relations via shallow graphical models, which is inadequate to model complicated human interactions.
In this paper, we propose a consistency-aware graph network, which combines the representative ability of  graph network and the consistency-aware reasoning to facilitate the HIU task.  Our network consists of three components, a backbone CNN to extract image features, a factor graph network to learn third-order interactive relations among participants, and a consistency-aware reasoning module to enforce labeling and grouping consistencies. Our key observation is that the  consistency-aware-reasoning bias for HIU can be embedded into an energy function, minimizing which  delivers consistent predictions. An efficient mean-field inference algorithm is proposed, such that all modules of our network could be trained jointly in an end-to-end manner. Experimental results show that our approach achieves leading performance on three  benchmarks.
\end{abstract}

\section{Introduction}
\label{sec:introduction}

Analyzing  human activities in natural scenes is a fundamental task to many potential applications like video surveillance \cite{2018Real}, key-event retrieval \cite{2017ER3}, social behavior interpretation \cite{2017Social} and sports analysis \cite{2018stagNet}. Abundant techniques have been developed for human activity recognition (HAR, where the goal is to assign an activity label to each image or video) \cite{choi2011learning,patron:2012structured,ji20133d,Annane2014Two,KongInteractive,wu2019learning,wu2019learning,2020empowerRN}, which have gained impressive progress on recognition accuracy. However, the task of human interaction understanding (HIU) is much less successful mainly because current methods learn human interactive relations via shallow graphical representations \cite{WangICCV2019,Wang2016A,wang2018understanding,patron:2012structured,choi2011learning,Yu2012Learning}, which is inadequate to model complicated human interactions, \eg fighting and chasing as two concurrent activities happening in the same scene. 

\begin{figure}
	\centering
	\includegraphics[width=.48\textwidth]{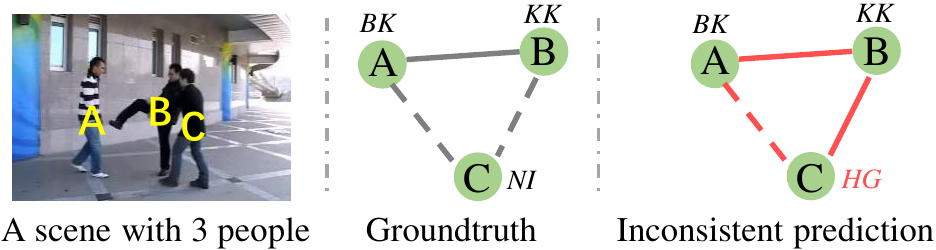}
	\caption{The graphical representation of HIU in a scene with three people. We decompose HIU into two sub-tasks: recognizing  person-wise actions (as denoted by the node labels, with KK, BK, HG, NI indicating \emph{kick, be-kicked, hug, no-interaction}, respectively) and predicting if any pair of people are interacting (solid edges) or not (dashed edges). Applying consistency-unaware models to such cases can lead to inconsistent predictions as highlighted by the red edges and labels (see Section~\ref{sec:introduction} for details). We address such issue by presenting a consistency-aware graph network with two types of third-order dependencies incorporated.}
	\label{fig:teaser_banner}
\end{figure}

As commonly done in literature \cite{patron:2012structured,WangICCV2019,wang2018understanding,Wang2016A}, we decompose HIU into  two sub-tasks illustrated by Figure~\ref{fig:teaser_banner} middle: 1) The individual action prediction task assigning each participant an action label; 2) The pairwise interactive prediction task determining if any pair of participants are interacting or not. Solving the two sub-tasks provides a way to disentangle concurrent human activities with multiple participants, as well as a comprehensive understanding to surveillance scenes. Though HIU performance had been lifted a lot by a conjunctive usage of deep features and rich contextual information, there still exist two main challenges. Since most existing works perform piece-wise learning of deep feature representations and contextual models \cite{WangICCV2019,wang2018understanding}, the first challenge is how to learn deep features and contextual relations jointly. The second challenge is how to ensure prediction consistency  for the two sub-tasks of HIU. In this paper, we tackle two types of prediction inconsistencies illustrated by Figure~\ref{fig:teaser_banner} right. The first type is called \emph{the labeling inconsistency}, \eg the action label of B (\ie \emph{kick}) is inconsistent with the action label of C (\ie \emph{hug}) as they are interacting (denoted by a solid edge). The second type is called \emph{the grouping inconsistency}, under the assumption that interacting people belong to the same group while non-interacting ones belong to separate groups, and vice versa. Consequently, the prediction \emph{(A, C) are not interacting} (denoted by the dashed edge) is inconsistent with the prediction that \emph{(A, B) are interacting} and \emph{(B, C) are interacting as well}.  To address the two challenges, we present a consistency-aware graph network (CAGNet), which consists of a backbone CNN to extract image features, a third-order graph network (TOGN) to learn human interactive context, and a consistency-aware reasoning (CAR) module to improve the consistency within action and interaction predictions. All components of CAGNet could be trained jointly and efficiently with GPU acceleration. We empirically validate the effectiveness of these three components on three benchmarks of human interaction understanding.

Our contributions are of three aspects. First, we propose a TOGN for HIU, which is more powerful than the widely adopted pairwise graph networks in terms of representing the interactive relations among people.  Second, we present an efficient CAR module to resolve the labeling and grouping inconsistencies within HIU predictions. Third, our proposed CAGNet, which takes the TOGN and CAR modules as its building-blocks, outperforms the state-of-the-art results by salient margins on three evaluated benchmarks.

\section{Related Work}
\label{sec:related work}

\textbf{Human Action/Activity Recognition} Since the invention of the two-stream network \cite{Annane2014Two},  numerous works on HAR (predicting each image or video an action class) have been proposed \cite{ji20133d,wang2015action,adascan,wang2016temporal,2017I3D,li2019actional,yan2019pa3d} in order to extract powerful feature representations of human motions. These approaches are also applicable to the recognition of collective activities wherein a number of participants perform a group activity. Nevertheless, an increasing number of works justify the importance of modeling the spatio-temporal correlations among action variables of different people \cite{choi2011learning,lan2012discriminative,2014WongUnderstanding,2016StructureInferenceM,2017Social,2017GernShu,2018Mostafa,2018stagNet,wu2019learning,2020empowerRN}. Early works in this vein explore conditional random fields (CRFs) \cite{choi2011learning,lan2012discriminative,2014WongUnderstanding}, while recent efforts contribute most on the joint learning of image features and human relations with RNN \cite{2016StructureInferenceM,2017Social,2017GernShu,2018stagNet,shu2019hierarchical} or deep graphical models \cite{2018Mostafa,wu2019learning,2020empowerRN}. These approaches are designed to predict each input an activity category, leaving the HIU task rather unsolved.

\textbf{Human Interaction Understanding} To understand human interactions, abundant conditional random field (CRF)-based models have been proposed \cite{Yu2012Learning,kong2015close,KongInteractive,patron:2010high,patron:2012structured,Wang2016A,wang2018understanding,WangICCV2019} to model the interactive relations in both spatial and temporal domains. The main drawback is that these CRFs are of shallow graphical-representations, which is neither effective in terms of learning complicated human interactions nor efficient in solving the associated maximum a posteriori inference \cite{WangICCV2019}. Moreover, they perform deep feature learning and relational reasoning separately, which typically results in sub-optimal solutions. Our CAGNet addresses these issues by presenting  a deep graph network, which synthesizes the feature-learning ability of CNNs and the contextual-modeling power of graphical representation.

\textbf{Graph Networks} have become popular choices to many tasks involving modeling and reasoning relations among components within a system \cite{battaglia2018relational,kipf2018neural,zhenZhangICCV19,guoShengICCV19,liFengICCV19,ZhangNIPS2020}. They share the computational efficiency of deep architectures while are more powerful and flexible in terms of modeling relations in non-grid structures, for instance, the correspondences between two sets of points in a matching problem \cite{zhenZhangICCV19}, the correlations between query and support pixels in one-shot semantic segmentation \cite{guoShengICCV19}, human gaze communication \cite{liFengICCV19}, and the  inter-person relations for collective activity classification \cite{wu2019learning}. As these networks operate on a graph structure, they are only able to capture pairwise relations. Very recently, work \cite{ZhangNIPS2020} proposes a factor graph neural network (FGNN) that enables the incorporation of high-order dependencies. Inspired by this, we propose the TOGN which shares the same feature-updating mechanism (detailed in Section~\ref{sec:prel}) with FGNN but uses customized third-order factor graphs to model the interactive relations in human activities.

\textbf{Deep Logical Reasoning} As a way to higher-level intelligence, logical reasoning has seen a renaissance in very recent years \cite{dong2018neural,yulin2020compo}. Since traditional logical reasoning has relied on methods and tools which are very different from deep learning models, such as Prolog language, SMT solvers and discrete algorithms, a key problem is how to bridge logic and deep models effectively and efficiently. Recent works viewed graph networks as a general tool to make such a connection.  For example, \cite{barcelo2019logical,battaglia2018relational} take graph networks to incorporate explicitly logic reasoning bias, \cite{mao2019neuro} builds a neuro-symbolic reasoning module to connect scene representation and symbolic programs, and work \cite{amizadeh2020neuro} introduces a differentiable first-order logic formalism for visual question answering. Like \cite{barcelo2019logical,battaglia2018relational}, our proposed CAR module  explicitly incorporates the consistency-aware-reasoning bias of HIU as well,  but accomplishes the reasoning differently via solving a particular energy minimization task.

\begin{figure*}[t]
	\centering
	\includegraphics[width=0.92\textwidth]{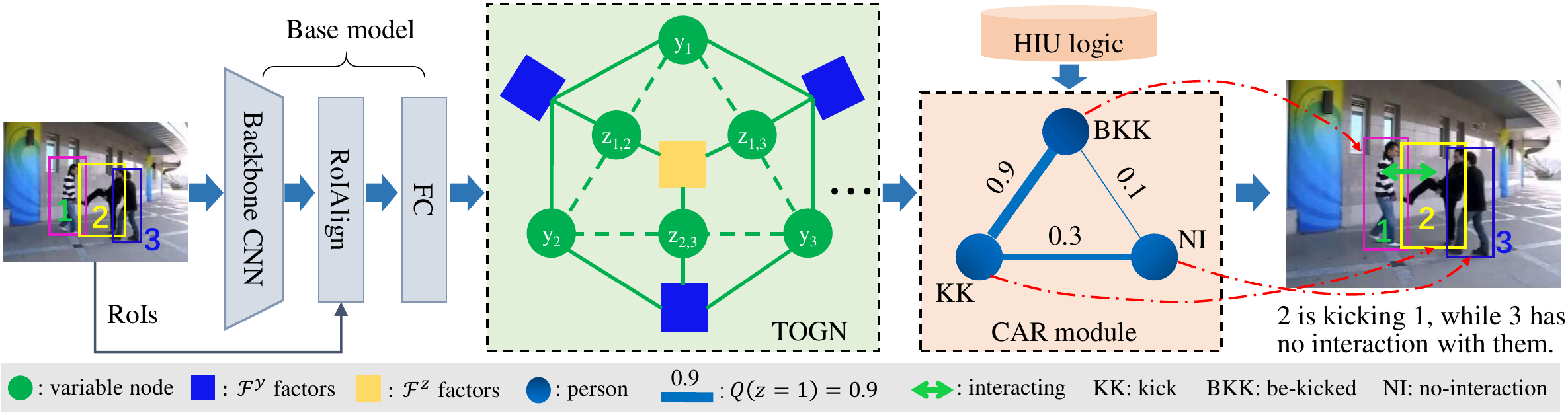}
	\caption{An overview of  the proposed CAGNet, which includes a base-model, a TOGN  and a CAR module.  The TOGN is designed to incorporate two types of factors to learn human-interaction-context, as indicated by yellow and blue nodes. Leveraging the consistency-aware-reasoning bias of HIU, our CAR block fixes possible inconsistent predictions and improves the interpretability of HIU. Here ``KK", ``BKK" and ``NI" represent ``kick", ``be-kicked" and ``no-interaction". All model parameters could be trained in an end-to-end manner.}
	\label{fig:overview}
\end{figure*}

\section{Preliminary}
\label{sec:prel}

As our TOGN shares the identical feature-updating mechanism with FGNN \cite{ZhangNIPS2020}, we first review this technique concisely. FGNN operates on a bipartite factor graph $\mathcal{G}=(\mathcal{V,C,E})$, where $\mathcal{V, C, E}$ denote the node set, the factor node set and the edge set respectively. Each $i\in \mathcal{V}$ is  associated with a discrete  variable $x_i\in \mathcal{X}_i$. Each edge $(c,i)\in \mathcal{E}$ connects a factor node $c\in \mathcal{C}$ and a node $i\in \mathcal{V}$. The factor graph defines a factorization of some function $f$ with $n$ variables. Specifically,  $f(x_1,\ldots,x_n)=\prod_{c\in C}f_c(\mathbf{x}_c)$, where $\mathbf{x}_c$ denotes the variables associated with the nodes which have edge-connections with $c$. In practice, the functions $f_c$ could be parameterized with deep networks.

Given $\mathcal{G}$, let $[\mathbf{f}_i^l]_{i\in \mathcal{V}}$ be a group of input node features, and let $[\mathbf{g}_c^l]_{c\in \mathcal{C}}$ be a group of input factor features, for the $l$-th layer of FGNN.  Let $[\mathbf{t}_e]_{e\in \mathcal{E}}$ be a group of edge features shared by all FGNN layers. Here $\mathbf{f}^l \in \mathbb{R}^{D_l}$, $\mathbf{g}^l \in \mathbb{R}^{D_l}$ and $\mathbf{t}\in \mathbb{R}^{H}$. FGNN  updates factor and node features separately via implementing two modules:
\begin{align}
\label{eq:mp_vf} &\mathbf{g}_c^{l+1} = \max_{i:(c,i)\in \mathcal{E}}\mathcal{Q}(\mathbf{t}_{ci}|\Phi_{VF}^{l})\mathcal{M}([\mathbf{g}_c^l,\mathbf{f}_i^l]|\Theta_{VF}^l),\\
\label{eq:mp_fv} &\mathbf{f}_i^{l+1} = \max_{c:(c,i)\in \mathcal{E}}\mathcal{Q}(\mathbf{t}_{ci}|\Phi_{FV}^l)\mathcal{M}([\mathbf{g}_c^l,\mathbf{f}_i^l]|\Theta_{FV}^l),
\end{align}
where $[\cdot,\cdot]$ denotes vector concatenation. The first equation is a factor-to-variable (FV) module and the second equation is a variable-to-factor (VF) module. $\mathcal{M}$ is a MLP (parameterized by $\Theta$ which is shared by all edges) maps the concatenation of factor and node features to a new feature vector of length $D_{l}$, and $\mathcal{Q}$ is another MLP (parameterized by $\Phi$, which is also shared by all edges) maps its input edge feature vector to  a $D_{l+1}\times D_l$ weight matrix. Here  $D_{l+1}$ denotes the length of the updated features (\ie the length of the input node features of the next layer), and the operator $\max$ actually performs max-pooling.  

Equations~\eqref{eq:mp_vf} and Equation~\eqref{eq:mp_fv} just comprise one layer of FGNN. To obtain a more powerful representation, one can stack a number of such layers, in which the output of the current layer is taken as the input to the subsequent layer.  We refer readers to \cite{ZhangNIPS2020} for more details of FGNN.


\section{Our Approach}
\label{sec:our_approach}

\textbf{Task Description and Notations~} Given an input image $\mathbf{I}$ and the bounding boxes (RoIs) of $n$ detected human bodies, the HIU task is decomposed into two sub-tasks: 1) predicting the action category $\mathbf{y} = (y_i)_{i=1}^n$ for every individual where $y \in \mathcal{Y}$ ($\mathcal{Y}$ takes all action categories), and 2) predicting all pairwise interactive relations $\mathbf{z}=(z_{j,k})_{j=1,\ldots, n;k=1,\ldots,n}$ for each pair of people, where $z_{j,k}\in \{0,1\}$ represents if the $j$-th and the $k$-th participants are interacting ($z_{s,t}=1$) or not $(z_{s,t}=0)$. All vectors in this paper will be column vectors unless otherwise stated.

\subsection{Model Overview}
\label{sec:formulation}

Figure~\ref{fig:overview} gives an overview of the proposed CAGNet, which consists of three components including a base-model, a TOGN and a CAR module. Given an input image and the detected human bodies as RoIs, the base-model takes a backbone CNN to extract features from the input, which are then processed by a RoIAlign module \cite{maskRCNNICCV2017} to generate local features for each individual. Afterwards the local features are  processed by one FC layer to generate \emph{base features} as inputs to TOGN. Our TOGN graph (Section~\ref{sec:fgnn}) includes two types of variable nodes (circles): one type is the $y$ node to represent the action category of the associated person, the other type is the $z$ node to represent the existence of interactive relation between a pair of people. The graph also includes a series of factor nodes (squares) in order to capture two types of third-order dependencies, respectively encoded by the $(y_i,y_j,z_{i,j})$ triplets (blue factor nodes) and the $(z_{u,v},z_{v,w},z_{u,w})$ triplets (the yellow factor node). We take the base features to initialize TOGN, and perform feature updating by passing messages between factor nodes and variable nodes such that rich contextual information could be embedded. Though TOGN is able to learn rich contextual representations to facilitate the HIU task, the labeling and grouping consistencies among variables are not explicitly modeled. To alleviate this, we introduce a CAR module, which essentially conducts a deductive reasoning leveraging the oracles presented in Section~\ref{sec:cfrm}. In practice the reasoning is implemented via solving a surrogate mean-field inference with differentiable high-order energy functions, which allows end-to-end learning of all modules within our CAGNet with GPU acceleration (Section~\ref{sec:end2end}).

\subsection{Third-Order Graph Network for HIU}
\label{sec:fgnn}

We now elaborate our TOGN for HIU in order capture two categories of third-order dependencies among action and interactive-relation variables.

\textbf{Third-Order Factor Graph} 
Formally, we define the factor graph as $\mathcal{G}=(\mathcal{V},\mathcal{F},\mathcal{E})$, where $\mathcal{V}$ is the set of variable nodes, $\mathcal{F}$ is the set of factor nodes, and $\mathcal{E}$ is the set of edges. The node set is split into two disjoint subsets: $\mathcal{V}=\mathcal{V}^{y} \cup \mathcal{V}^{z}, \mathcal{V}^{y} \cap \mathcal{V}^{z}=\emptyset$. Specifically, $\mathcal{V}^{y}=\{ 1,\cdots, n\}$, and $\mathcal{V}^{z}=\{n+1, \cdots, n+\tbinom{n}{2}\}$. For each node $i\in \mathcal{V}^y$, a variable $y_i\in \mathcal{Y}$ is associated with it to represent the action category of the $i$-th individual. Let $g(u,v)$ be a function:
\begin{align}
g(u,v):\mathcal{V}^{y} \times \mathcal{V}^{y} \mapsto \mathcal{V}^{z},  \forall u, v \in \mathcal{V}^y, u < v.
\end{align}
For each node  $k\in \mathcal{V}^z$, a variable $z_{u,v}\in\{0,1\}$ is associated with it to represent if the pair of people $(u,v)$ are interacting ($z_{u,v}=1$) or not ($z_{u,v}=0$), where $k=g(u,v)$.

To encode different relations, we create two groups of factor nodes. The first group is
\begin{align}
\mathcal{F}^{y}=\{ (i,\ j,\ g(i,j))\ |\ \forall i,j \in \mathcal{V}^{y},\ i < j\},
\end{align}
which is taken to implicitly model the correlations among  $y_i$, $y_j$ and $z_{i,j}$ based on their base features. Intuitively, action labels $(y_i,y_j)$ are highly correlated when the associated people are interacting (taking the \emph{kicking} interaction in Figure~\ref{fig:overview} as an example), while this correlation vanishes if they are not interacting (\eg, Person 2 and Person 3 in Figure~\ref{fig:overview}).

The second group of factors is defined as
\begin{align}
\mathcal{F}^{z}=\{ (g(r,s),g(s,t),g(r,t))\ |\ \forall r,s,t \in \mathcal{V}^{y},\ r<s<t \},
\end{align}
which is leveraged to implicitly model the correlations among $z_{r,s}$, $z_{s,t}$ and $z_{r,t}$ for each triplet of people $(r,s,t)$. With such factors,  we encourage the model to learn representations for the prediction of consistent interactive relations for each triplet. Fortunately, higher-order consistencies can be guaranteed if all third-order consistencies are satisfied (detailed in Section~\ref{sec:cfrm}).

In summary, the factor node set is $\mathcal{F}=\mathcal{F}^{y}\cup\mathcal{F}^{z}$. Given $\mathcal{F}$ and $\mathcal{V}$, the edge set $\mathcal{E}$ is set up by connecting variable nodes with factor nodes. Specifically, for each factor node $c=(i,j,k)\in \mathcal{F}$, we put three edges $(c,i)$, $(c,j)$ and $(c,k)$ into $\mathcal{E}$, which finalizes the construction of the TOGN graph.

\textbf{Initial Node Feature}
For each node $i\in \mathcal{V}^y$, let $\mathbf{\phi}_i$ be the \emph{base feature} extracted from the bounding box region of the $i$-th person using the base-model. For each $(u,v)\in \mathcal{V}^y\times \mathcal{V}^y$, $u<v$, let $j = g(u,v) \in \mathcal{V}^z$. We concatenate $\mathbf{\phi}_u$ and $\mathbf{\phi}_v$, and use the concatenation as the \emph{base feature} (denoted by $\mathbf{\phi}_j$) for the variable node $j$. In order to compute the initial node features, we apply to the \emph{base features} the linear transformations: 
\begin{align}
\mathbf{f}^1_i&={\rm FC}^y(\mathbf{\phi}_i),~\forall i \in \mathcal{V}^y,\\
\mathbf{f}^1_j&={\rm FC}^z(\mathbf{\phi}_j), ~\forall j \in \mathcal{V}^z,
\end{align}
which project the original features into $\mathbb{R}^{D_1}$ space:

\textbf{Initial Factor Feature}
The factor features are computed based on node features. For each factor node $c=(i,j,g(i,j)) \in \mathcal{F}^y$, the initial factor feature $\mathbf{g}^1_c \in \mathbb{R}^{D_1}$ is computed with:
\begin{align}
\mathbf{g}_c^1=\frac{\mathbf{f}_i^1+\mathbf{f}_j^1+\mathbf{f}_{g(i,j)}^1}{3}.
\end{align}

For each $d=(g(r,s),g(s,t),g(r,t)) \in \mathcal{F}^z$, the associated factor feature $\mathbf{g}_d^1 \in \mathbb{R}^{D_1}$ is obtained using:
\begin{align}
\mathbf{g}^1_d=\frac{\mathbf{f}_{g(r,s)}^1+\mathbf{f}_{g(s,t)}^1+\mathbf{f}_{g(r,t)}^1}{3}.
\end{align}

\textbf{Edge Feature}
For each edge $e=(q,p) \in \mathcal{E}$, the related feature $\mathbf{t}_e \in \mathbb{R}^{H}$ is given by:
\begin{align}
\mathbf{t}_e={\rm ReLU}\big({\rm FC}^e([\mathbf{f}_p^1,\mathbf{g}_q^1])\big),
\end{align}
where $p \in \mathcal{V}, q \in \mathcal{F}$ and ${\rm FC^e}$ maps the concatenated feature vector to $\mathbb{R}^{H}$ space.

Taking as inputs the factor graph and the initial features, the first TOGN layer performs feature updating with the method described in Section~\ref{sec:prel}. Afterwards we take the updated features as inputs to the next TOGN layer (which shares the factor graph and the feature updating algorithm with the first TOGN layer, but using different model parameters), and perform feature updating again. Empirically, we find that TOGN with 10 such layers works well for HIU.
Finally, we compute the classification scores for individual actions and pairwise interactive relations using
\begin{align}
\label{eq:fgnn_score_y}
\theta_i&= {\rm Softmax}\big(\alpha(\mathbf{f}_i^{*})\big)~ \forall i \in \mathcal{V}^y,\\
\label{eq:fgnn_score_z}
\theta_j&= {\rm Softmax}\big(\beta(\mathbf{f}_j^{*})\big)~ \forall j \in \mathcal{V}^z,
\end{align}
where $\mathbf{f}_i^*$, $\mathbf{f}_j^*$ are updated node features output by the last TOGN layer, $\alpha$ and $\beta$ are linear functions which compute classification scores for individual actions ($\theta_i\in \mathbb{R}^{|\mathcal{Y}|}$) and pairwise interactive relations ($\theta_j\in \{0,1\}$), respectively.

\subsection{Consistency-Aware Reasoning}
\label{sec:cfrm}
To resolve possible inconsistencies (recall Figure~\ref{fig:teaser_banner}) incurred by using consistency-unaware models like CNN, PGNN and TOGN, we first present two deductive reasoning bias for human interaction understanding:
\begin{itemize}
	\item \textbf{The compatibility oracle}: For any pair of interacting (denoted by $\leftrightarrow$) people $A$ and $B$, their action categories must be compatible (denoted by $\odot$). In logical words, this rule is represented by $A \leftrightarrow B \Rightarrow y_A \odot y_B$.
	\item \textbf{The transitivity oracle}: Considering the interactive relations among a triplet of people  $(A,B,C)$, we have $(A \leftrightarrow B) ~\&~ (B \leftrightarrow C) \Rightarrow (A \leftrightarrow C)$.
\end{itemize}

\begin{figure}[t!]
	\centering
	\includegraphics[width=.48\textwidth]{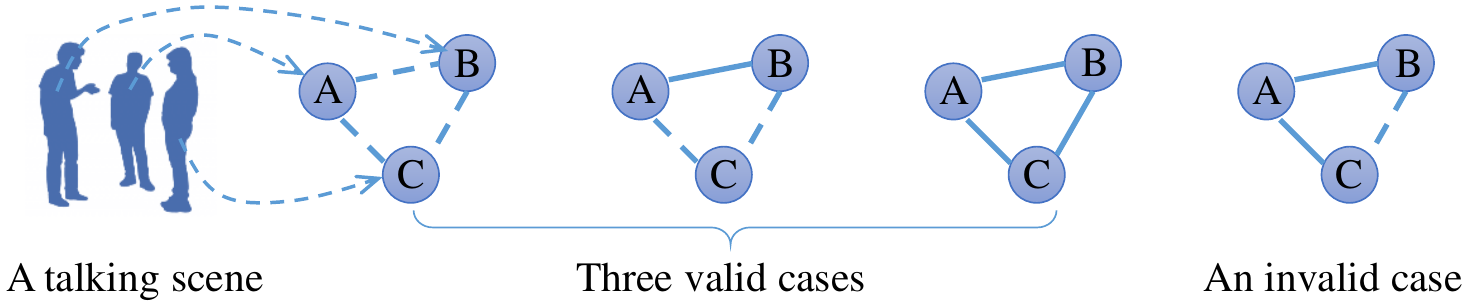}
	\caption{Predictions obey and violate the \emph{transitivity oracle} in a talking scene with three people. Here solid edges represent predicted interactive relations ($z=1$) and dashed edges indicate predicted non-interactive relations ($z=0$).}
	\label{fig:example_trans_comp}
\end{figure}

Typical compatible examples include (\emph{handshake, handshake}), (\emph{pass, receive}) and (\emph{punch, fall}), and typical incompatible examples are (\emph{handshake, hug}), (\emph{punch, pass}), (\emph{highfive, handshake}). Instead of predesignating such compatibility, which might change across datasets, our CAR module is able to learn them directly from data. Examples obey or violate the \emph{transitivity}  are shown in Figure~\ref{fig:example_trans_comp}. Though this oracle only considers triplets of people, it is straightforward to prove that  the higher-order transitivity associated with an arbitrary number of people is simply a conclusion of the third-order transitivity. Intuitively, by enforcing the transitivity across all triplets, participants in the scene are split into different groups, such that individuals in the identical group are interacting with each other, while people in different groups have no interaction.

With such oracles, predictions of the TOGN described in Section~\ref{sec:fgnn} could be refined by applying the traditional logical reasoning algorithms like resolution. However, embedding such reasoning into deep learning frameworks directly is highly challenging. As a workaround, our reasoning approach first embeds the knowledge into an energy function defined by
\begin{align}
E&(\mathbf{y},\mathbf{z};\mathbf{x})=\sum_{i\in \mathcal{V}^y} -\theta_i(y_i) + \sum_{(j,k,l)\in \mathcal{F}^y}\big[-\theta_{j,k}(z_{j,k}) + \notag\\
&K^{\mathcal{C}}(y_j,y_k,z_{j,k}) \big]+\sum_{(r,s,t)\in \mathcal{F}^z}K^{\mathcal{T}}(z_{r,s},z_{s,t},z_{r,t}),
\label{eq:energy_function}
\end{align}
where $\theta_{j,k}(z_{j,k})=\theta_{g(j,k)}(z_{j,k})$. The data terms  $-\theta_i$ and $-\theta_{j,k}$ (computed by Equations~\eqref{eq:fgnn_score_y} and \eqref{eq:fgnn_score_z}) are utilized to penalize particular $y$-label and $z$-label assignments  respectively based on the learned deep representations. The functions $K^{\mathcal{C}}$ and $K^{\mathcal{T}}$ are the so-called $P^n$-Potts models \cite{kohli2007p3} defined by

\begin{align}
K^{\mathcal{C}}(y_j,y_k,z_{j,k})= 
\begin{cases}
\lambda^{\mathcal{C}}(y_j,y_k) & {\rm if~} z_{j,k}=1,\\
0 & {\rm otherwise.}
\end{cases}
\end{align}

\begin{align}
K^{\mathcal{T}}(z_{r,s},z_{s,t},z_{r,t})= 
\begin{cases}
\lambda^{\mathcal{T}} & \!\!\!\!{\rm ~if~} (z_{r,s},z_{s,t},z_{r,t})\in \Gamma,\\
0 & {\rm~otherwise.}
\end{cases}
\end{align}
Here $\Gamma$ is a set $\{(1,1,0), (1,0,1),(0,1,1)\}$ that includes all cases violating the transitivity oracle, $\lambda^{\mathcal{C}}(y_j,y_k)$ and $\lambda^{\mathcal{T}}$ are penalties incurred by predictions which violate the compatibility and transitivity oracles. It is easy to check that when $\lambda^{\mathcal{C}}$ and $\lambda^{\mathcal{T}}$ are sufficiently large, minimizing the energy \eqref{eq:energy_function} delivers desirable $\mathbf{y}$ and $\mathbf{z}$ predictions which satisfy the \emph{compatibility} and \emph{transitivity} oracles. In this paper, instead of predesignating suitable $\lambda^{\mathcal{C}}$ and $\lambda^{\mathcal{T}}$ values, we learn them from  data  in conjunction with other parameters of  CAGNet.

\textbf{Mean-Field Inference} Minimizing \eqref{eq:energy_function} is NP-complete. Here we derive an efficient mean-field inference algorithm by first approximating the joint distribution $P(\mathbf{y},\mathbf{z}|\mathbf{x})\propto \exp(-E(\mathbf{y},\mathbf{z};\mathbf{x}))$ with a product of independent marginal distributions: 
\begin{align}
P(\mathbf{y},\mathbf{z}|\mathbf{x}) \approx \prod_{i\in \mathcal{V}^y} Q_i(y_i)\prod_{l\in \mathcal{V}^z:g(j,k)=l} Q_{j,k}(z_{j,k}).
\end{align}

\begin{table*}[t!]
	\centering
	\small
	\begin{tabular}{l|ccc|ccc|ccc}
		\hline
		& & UT & & & BIT & & & TVHI & \\
		Method & F1 & Accuracy & mean IoU & F1 & Accuracy &  mean IoU & F1 & Accuracy & mean IoU\\
		\hline
		\hline
		VGG19  \cite{Simonyan15} & 85.69 & 91.68 & 69.03 & 85.22 & 89.60 & 67.03  & 70.68 & 76.90 & 52.30\\
		ResNet50 \cite{He2016Deep} & 90.62 & 94.64 & 76.70 & 87.12 & 91.20 & 71.41  & 81.18 & 82.61 & 66.33\\
		Inception V3 \cite{szegedy2016rethinking} & 92.20 & 95.86 & 80.30 & 87.84 & 91.61 & 72.00  & 83.00 & 86.91 & 71.53 \\
		
		Base model + CAR  & 92.81 & 96.26 & 81.51 & 88.72 & 92.23 & 73.99  & 83.07 & 87.23 & 72.29 \\
		
		TOGN (ours) &93.45 &96.50 &84.53 &91.26 &94.84 &78.27 &90.41 &92.51 &79.40 \\
		TOGN+${\rm CAR^C}$ (ours) &94.32 &97.03 &85.06 &92.70 &95.34 &80.78 &91.90 &93.44 &82.07 \\
		TOGN+${\rm CAR^T}$ (ours) &93.82 &96.71 &83.90 &92.30 &95.20 &80.22 &90.35 &92.63 &79.05 \\
		TOGN+${\rm CAR^{CT}}$ (ours) &\textbf{94.55} &\textbf{97.06} &\textbf{85.50} &\textbf{92.79} &\textbf{95.41} &\textbf{81.32} &\textbf{92.83} &\textbf{95.29} &\textbf{84.02}\\
		\hline
	\end{tabular}
	\caption{
		Ablation study on three benchmarks. All results are in percentage. The proposed TOGN performs much better than the best base model (Inception V3). The proposed CAR module further improves  HIU results  by clear margins. Bold texts denote best results.}
	\label{tbl:abstudy}
\end{table*}

Then we derive the mean-field updates of all marginal distributions using the techniques described in \cite{vineet2014filter}, which gives
\begin{align}
&\tilde{Q}^t_i(y_i)=\!\!\!\!
\sum_{j\in V\setminus\{i\}}\!\!\sum_{y_j} \lambda^{\mathcal{C}}(y_i,y_k) Q^{t-1}_j(y_j) Q^{t-1}_{i,j}(z_{i,j}=1),\label{eq:action_update}\\
&Q^t_i(y_i)=\frac{\exp\big(\theta_i(y_i)-\tilde{Q}^t_i(y_i)\big)}{Z_i},
\label{eq:action_update_norm}
\end{align}
where $Z_i$ is a normalization constant. The marginal distributions on $z$ variables are
\begin{align}
&\tilde{Q}^t_{k,l}(z_{k,l}) = \sum_{y_k,y_l} z_{k,l} \lambda^{\mathcal{C}}(y_k,y_l) Q^{t-1}_k(y_k)Q^{t-1}_l(y_l) +\notag\\
&\quad\quad\quad\quad\sum_{m \in V\setminus\{k,l\}}
\sum_{z_{k,m},z_{m,l}}\mathbbm{1}\big((z_{k,m},z_{m,l},z_{k,l})\in \Gamma\big) \lambda^{\mathcal{T}} \notag\\
&\quad\quad\quad\quad\quad Q^{t-1}_{k,m}(z_{k,m}) Q^{t-1}_{m,l}(z_{m,l}),\label{eq:inter_update}\\
&Q^t_{k,l}(z_{k,l})=\frac{\exp\big(\theta_{k,l}(z_{k,l})-\tilde{Q}^t_{k,l}(z_{k,l})\big)}{Z_{k,l}},
\label{eq:inter_update_norm}
\end{align}
where $\mathbbm{1}(\cdot)$ is an indicator function (gives $1$ if the testing condition holds, and $0$ otherwise), $t\in \{1,2,\ldots,T\}$,  $Z_{k,l}$ is a normalization constant. We initialize the marginal distributions $Q_i^0(y_i)$, $Q_{k,l}^0(z_{k,l})$ by applying the {softmax} function to the scores output by the graph network. The inference is summarized by Algorithm~\ref{alg:meanfield}. Note that we can perform the updates of all expectations (Equation~\eqref{eq:action_update} and \eqref{eq:inter_update}) and marginal probabilities (Equation~\eqref{eq:action_update_norm} and \eqref{eq:inter_update_norm}) in parallel, which yields very efficient inference.

\begin{algorithm}[t!]
	\caption{The mean-field inference.}
	\label{alg:meanfield}
	\KwIn{The graph $\mathcal{G}$, $\theta_i(y_i)$,  $\theta_{k,l}(z_{k,l})$, $\lambda^{\mathcal{C}}$ and $\lambda^{\mathcal{T}}$.}
	\KwOut{$\check{\theta}_i(y_i)$, $\check{\theta}_{k,l}(z_{k,l})$.}
	\textbf{Initialization}: Let $Q_i^0(y_i)=\frac{\exp(\theta_i(y_i))}{Z_i}$, and let $Q_{k,l}^0(z_{k,l})=\frac{\exp(\theta_{k,l}(z_{k,l}))}{Z_{k,l}}$.\\
	\For{$t=1,2,\ldots,T$}
	{
		Compute $\tilde{Q}^t_i(y_i)$, $\tilde{Q}^t_{k,l}(z_{k,l})$, ${Q}^t_i(y_i)$ and ${Q}^t_{k,l}(z_{k,l})$ using Equations~\eqref{eq:action_update} to \eqref{eq:inter_update_norm}.
	}
	$\check{\theta}_i(y_i) \leftarrow \theta_i(y_i)-\tilde{Q}^T_i(y_i)$, 
	$\check{\theta}_{k,l}(z_{k,l}) \leftarrow \theta_{k,l}(z_{k,l})-\tilde{Q}^t_{k,l}(z_{k,l})$.
\end{algorithm}

As mentioned, Algorithm~\ref{alg:meanfield} is a surrogate of the consistency-aware reasoning task taking the two oracles as its knowledge-base. This algorithm actually forms the last layer of our CAGNet, which outputs updated action scores $\check{\theta}_i~\forall i \in V^y$ and  interactive scores $\check{\theta}_{j,k}$, $\forall l \in \mathcal{V}^z$ and $g(j,k)=l$. Our experimental results in Section~\ref{sec:experiment} demonstrate that such updated scores are able to deliver more consistent HIU predictions.

\subsection{End-to-End Learning}
\label{sec:end2end}
The mean-field inference algorithm allows the back-propagation of the error signals $\frac{\partial{Loss}}{\partial{Q}}$ to all parameters of CAGNet (including that of the base-model, the TOGN and the CAR module), which enables the joint training of all parameters from scratch. In practice, we resort to a two-stage training due to the limitation of computational resources. The first stage learns a base-model with the backbone CNN initialized by a  model pre-trained on ImageNet. The second stage trains the TOGN, $\lambda^{C}(y_j,y_k)$ and $\lambda^{T}$ jointly with fixed backbone-parameters. We train all models using the identical cross-entropy losses computed on both $\mathbf{y}$ and $\mathbf{z}$ predictions.

\textbf{Implementation Details} Our implementation is based on PyTorch deep learning toolbox and a workstation with  three pieces of NVIDIA GeForce GTX 1080 Ti GPU. We test several backbone CNNs  including VGG19 \cite{Simonyan15}, ResNet 50 \cite{He2016Deep} and Inception V3 \cite{szegedy2016rethinking}. We use the official implementation of RoIAlign by PyTorch, which outputs feature maps with a size of $5\times 5 \times 1056$ (using Inception V3).  We add dropout (the ratio is 0.3) followed by a layer-normalization to every FC layer of CAGNet except for the ones computing final classification scores.  For the mean-field inference we set $\lambda^{\mathcal{C}}(y_j,y_k)=0.5$ and $\lambda^{\mathcal{T}}=0.1$ for initialization.  We adopt mini-batch SGD with Adam to learn the network parameters, and train all models in 200 epochs. We augment training data with random combinations of scaling, cropping,  horizontal flipping and color jittering. For the scaling and flipping operations, the bounding boxes are scaled and flipped as well.

\section{Experiment}
\label{sec:experiment}
\textbf{Dataset} We use three benchmarks including  UT \cite{UT-Interaction-Data}, BIT \cite{Yu2012Learning} and TVHI \cite{patron:2010high}. UT contains 120 short videos of 6 action classes: \emph{handshake}, \emph{hug}, \emph{kick}, \emph{punch}, \emph{push} and \emph{no-action}. As done by \cite{wang2018understanding}, we extend original action classes by introducing a passive class for each of the three asymmetrical action classes including kick, punch and push (\emph{be-kicked}, \emph{be-punched} and \emph{be-pushed}). Consequently, we have 9 action classes in total. Following \cite{wang2018understanding}, we split samples of UT into 2 subsets for training and testing. BIT covers 9  interaction classes including \emph{box}, \emph{handshake}, \emph{highfive}, \emph{hug}, \emph{kick}, \emph{pat}, \emph{bend}, \emph{push} and \emph{others}, where each class contains 50 short videos. Of each class 34 videos are chosen for training and the rest for testing as  recommended by \cite{Yu2012Learning}. TVHI contains 300 short videos of television shows, which covers 5 action classes including \emph{handshake}, \emph{highve}, \emph{hug}, \emph{kiss} and \emph{no-action}. As suggested by  \cite{patron:2010high}, we split samples of TVHI into two parts for training and testing.

\subsection{Ablation}
\label{sec:abstudy}

\textbf{Evaluation Metric} Since the numbers of instances across different classes are significantly imbalanced, we use multiple metrics including \emph{F1-score}, \emph{overall accuracy} and  \emph{mean IoU} for evaluation. Specifically, we calculate the \emph{macro-averaged-F1} scores on $\mathbf{y}$ and $\mathbf{z}$ predictions respectively (using the \emph{f1\_score} function in \emph{sklearn} package), and present the mean of the two F1 scores. Likewise,  \emph{overall accuracy} calculates the mean of the action-classification accuracy and the interactive-relation-classification accuracy. To obtain \emph{mean IoU}, we first compute IoU value on each class, then average all IoU values. We first analyze the capabilities of different components in the proposed CAGNet, using results provided by Table~\ref{tbl:abstudy}.

\textbf{Choice of Backbone-CNN.} Here we evaluate base models (see Figure~\ref{fig:overview}) taking different backbone CNNs to extract image features. We test three popular backbones: VGG19 \cite{Simonyan15}, Inception V3 \cite{szegedy2016rethinking} and ResNet50 \cite{He2016Deep}, and the results correspond to the first three rows (from top to bottom) in Table~\ref{tbl:abstudy}. Inception V3 performs notably better than other backbones on all benchmarks. The reason might be that Inception V3 is able to learn multi-scale feature representations, which stacks into a feature pyramid to better capture the appearance of human actions. Hence we use Inception V3  as the backbone for all subsequent experiments.


\textbf{Power of  the TOGN}
Remember that the proposed TOGN takes the features extracted by Inception V3 as inputs, and learn third-order dependencies among structured variables. Overall, the proposed TOGN outperforms all basemodels on the three benchmarks under all considered metrics. In particular, TOGN results are moderately higher than the best basemodel (\ie Inception V3) on UT, and are significantly better than the basemodel on BIT (91.26 \vs 87.84 on F1) and TVHI (90.41 \vs 83.0 on F1), thanks to the rich contextual representations learned by our TOGN. Note that the performance gain on UT is much less compared with results on other benchmarks. This is probably because human interactions in UT (each video contains just two individuals, and the background is clear) are simpler than BIT and TVHI, and the performance on this dataset tends to be saturated.

\textbf{Effect of the CAR Module} Here we compare four models: 1) \emph{Base model + CAR} that consists of the base-model (Inception V3 as backbone) followed by the proposed CAR module; 2) TOGN+${\rm CAR^C}$ is the proposed CAGNet without taking the \emph{transitivity oracle} into consideration; 3) TOGN+${\rm CAR^T}$ is the proposed CAGNet without taking the \emph{compatibility oracle} into consideration; 4) TOGN+${\rm CAR^{CT}}$ actually is our full CAGNet. Here we can draw two conclusions based on the results in Table~\ref{tbl:abstudy}. First, the incorporation of the CAR module (Base model + CAR and TOGN+${\rm CAR^{CT}}$) boosts HIU performance (2.42, 2.78 and 4.62 points better than TOGN on TVHI in terms of F1, Accuracy and mean IoU), which validates the significance of exploiting consistency-aware-reasoning. Second, both oracles are critical to achieve the best results. Though incorporating either the compatibility (TOGN+${\rm CAR^C}$) or the transitivity oracle (TOGN+${\rm CAR^T}$) already performs better than TOGN, TOGN with both oracles  (TOGN+${\rm CAR^{CT}}$) performs notably  better than using each of them separately, which suggests that these two oracles complement each other for HIU.

\begin{table}[t!]
	\centering
	\small
	\begin{tabular}{l|ccc}
		\hline
		Method & F1 (\%) & Accuracy (\%) & mean IoU (\%)\\
		\hline
		\hline
		GN \cite{wu2019learning} & 80.57 & 82.76  & 66.82  \\
		Modified GN \cite{wu2019learning}  & 84.18 & 87.86 & 71.31 \\
		Joint + AS \cite{wang2018understanding} & 83.50 & 87.33 & 71.64 \\
		QP + CCCP \cite{WangICCV2019} & 83.42 & 87.25 & 71.61\\
		CAGNet (ours) &\textbf{92.83} &\textbf{95.29} &\textbf{84.02} \\
		\hline
	\end{tabular}
	\caption{Comparison with recent methods on TVHI. Our CAGNet overshoots competitive models under all evaluation metrics.}
	\label{tbl:sota_tvhi}
\end{table}

\begin{table}[t!]
	\centering
	\small
	\begin{tabular}{l|ccc}
		\hline
		Method & F1 (\%) & Accuracy (\%) & mean IoU (\%)\\
		\hline
		\hline
		GN \cite{wu2019learning} & 70.52 & 78.52 & 65.89 \\
		Modified GN \cite{wu2019learning}  &  89.95 & 93.38 & 76.42 \\
		Joint + AS \cite{wang2018understanding} & 88.61 & 91.77 & 72.12\\
		QP + CCCP \cite{WangICCV2019} & 88.80 &  91.92 & 72.46\\
		CAGNet (ours) &\textbf{92.79} &\textbf{95.41} &\textbf{81.32} \\
		\hline
	\end{tabular}
	\caption{Comparison with recent methods on BIT. Our CAGNet performs much better than other recent approaches.}
	\label{tbl:sota_bit}
\end{table}

\begin{table}[t!]
	\centering
	\small
	\begin{tabular}{l|ccc}
		\hline
		Method & F1 (\%) & Accuracy (\%) & mean IoU (\%)\\
		\hline
		\hline
		GN \cite{wu2019learning} & 89.25 & 91.78 & 78.24  \\
		Modified GN \cite{wu2019learning} & 93.38 & 96.39 &  84.13 \\
		Joint + AS \cite{wang2018understanding} & 92.20 & 95.86 & 80.30 \\
		QP + CCCP \cite{WangICCV2019}  & 89.71 & 93.23 & 80.35\\
		CAGNet (ours) &\textbf{94.55} &\textbf{97.06} &\textbf{85.50} \\
		\hline
	\end{tabular}
	\caption{Comparison with recent methods on UT. Our CAGNet performs moderately better than other recent approaches.}
	\label{tbl:sota_ut}
\end{table}


\begin{figure*}[t!]
	\centering
	\subfigure[Groundtruth]{
		\begin{minipage}[b]{0.24\linewidth}
			\includegraphics[width=1\linewidth]{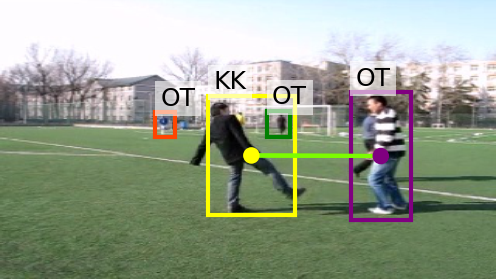}
    		\includegraphics[width=1\linewidth]{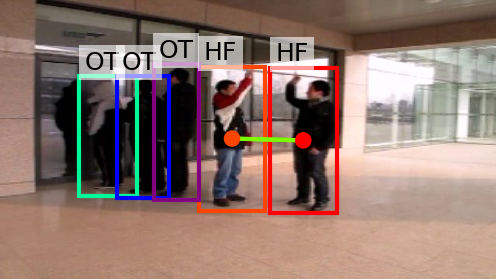}
			\includegraphics[width=1\linewidth]{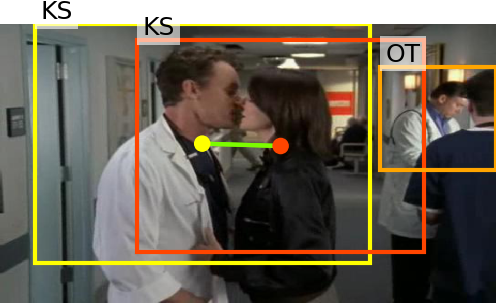}
			\includegraphics[width=1\linewidth]{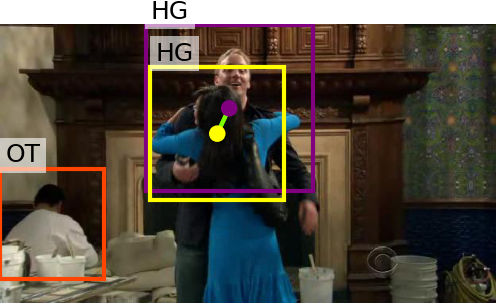}
	\end{minipage}}
	\subfigure[Base model]{
		\begin{minipage}[b]{0.24\linewidth}
			\includegraphics[width=1\linewidth]{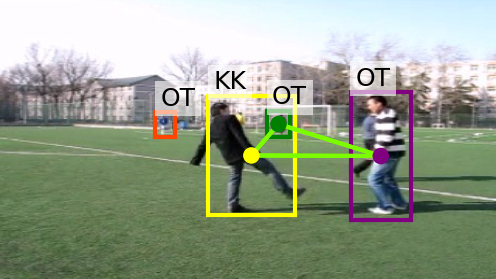}
			\includegraphics[width=1\linewidth]{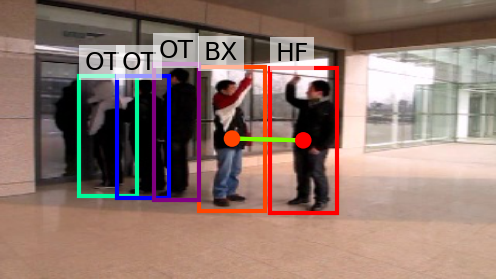}
			\includegraphics[width=1\linewidth]{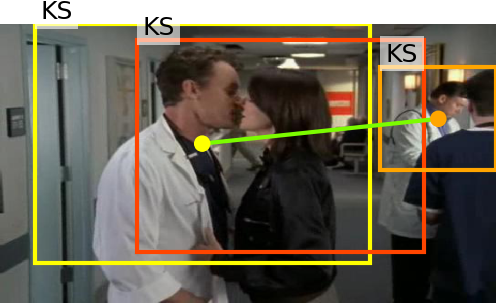}
			\includegraphics[width=1\linewidth]{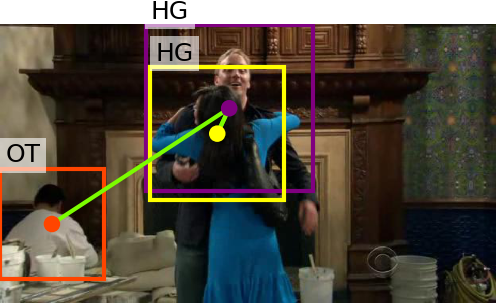}
	\end{minipage}}
	\subfigure[Modified GN]{
		\begin{minipage}[b]{0.24\linewidth}
			\includegraphics[width=1\linewidth]{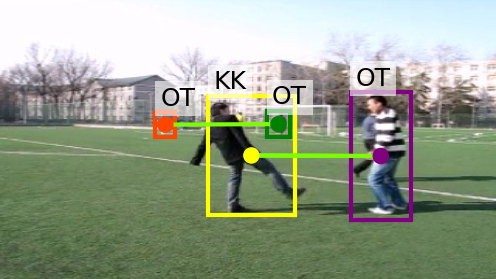}
			\includegraphics[width=1\linewidth]{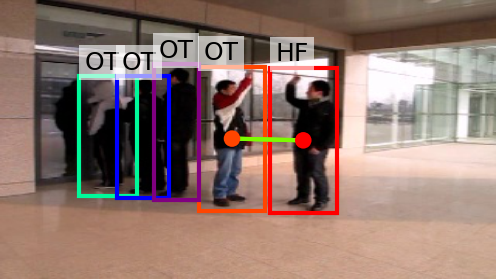}
			\includegraphics[width=1\linewidth]{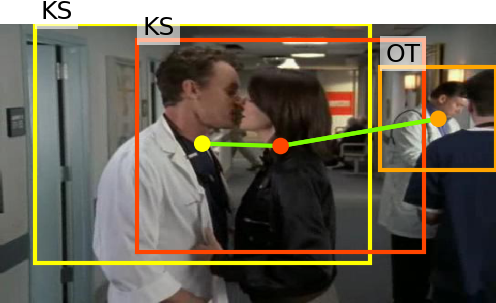}
			\includegraphics[width=1\linewidth]{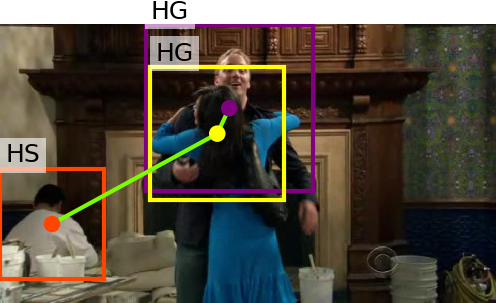}
	\end{minipage}}
	\subfigure[CAGNet]{
		\begin{minipage}[b]{0.24\linewidth}
			\includegraphics[width=1\linewidth]{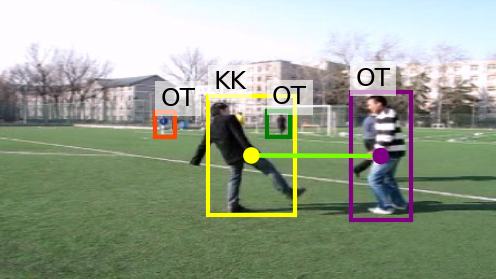}
			\includegraphics[width=1\linewidth]{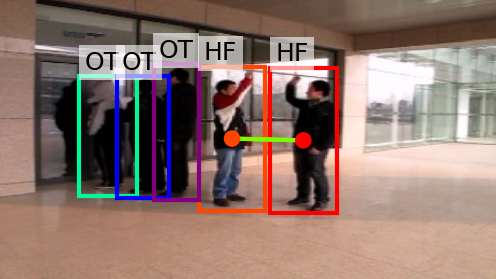}
			\includegraphics[width=1\linewidth]{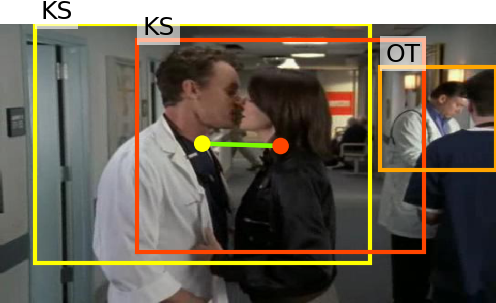}
			\includegraphics[width=1\linewidth]{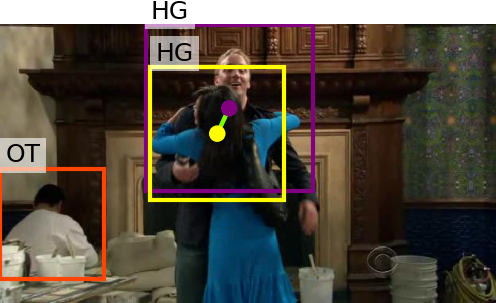}
	\end{minipage}}
	\caption[*]{
		Visualize HIU results predicted by different models. Figures are best viewed in color. Each row shows an example (from top to bottom, the first two rows are from BIT and the rest are from TVHI). Columns from left to right correspond to results of \emph{groundtruth}, \emph{base-model}, \emph{Modified GN}, and \emph{CAGNet}. Green lines denote predicted interactive pairs ($z=1$). Texts present predicted individual actions (y variables), where \emph{HG, KS, HF, KK, OT} mean \emph{hug, kiss, high-five, kick, others} respectively. Note that the predictions of CAGNet (the rightmost column) always obey the two oracles defined in Section~\ref{sec:cfrm}.
	}
	\label{fig:hiu_vis_result}
\end{figure*}

\subsection{Comparison with Recent Methods}
We consider three recent approaches. \textbf{\emph{Joint + AS}} \cite{wang2018understanding}  first extracts motion features of individual actions with backbone CNN. Afterwards the deep and contextual features of human interactions are fused by structured SVM. This method is able to predict $\mathbf{y}$ and $\mathbf{z}$ in a joint manner. \textbf{\emph{QP + CCCP}} \cite{WangICCV2019} takes a structured model to represent the correlations between $\mathbf{y}$ and $\mathbf{z}$ variables as well. It also developed an inference algorithm (namely \emph{QP + CCCP}) to solve the related inference problem. 
\textbf{\emph{GN}} \cite{wu2019learning} is a recent state-of-the-art for recognizing collective human activities. This model is empowered by both the representative ability of deep CNNs and the attention mechanism of PGNN. Note that \emph{GN} does not yield $\mathbf{z}$ predictions. We fix this with two solutions. First, we just set $z_{i,j}=1$ if the learned relation value is greater than $0.5$ (see Equation~(2) in \cite{wu2019learning}), otherwise we set $z_{i,j}=0$ (this solution does not introduce new parameters). Second, we attach a head to the tail of \emph{GN} to make $z$ predictions, and train parameters of this head with cross-entropy loss. We call such a solution the \textbf{\emph{Modified GN}}. We find that such a
straightforward modification offers a boost of performance on HIU compared against the original GN (see Table~\ref{tbl:sota_tvhi} to Table~\ref{tbl:sota_ut}). The reason is that \emph{Modified GN} is trained with additional supervision on interactive relations, which guides the network to learn more useful representations for the prediction of $z$.

For fair comparison, all methods take Inception V3 as the backbone to extract image features.
Results on three datasets are provided in Table~\ref{tbl:sota_tvhi}, Table~\ref{tbl:sota_bit} and  Table~\ref{tbl:sota_ut}. We can see that CAGNet outperforms \emph{modified GN} and shallow structured models ( \emph{Joint + AS} and \emph{QP + CCCP})  significantly on all evaluated benchmarks. Compared with CAGNet which is able to model higher-order relations, \emph{modified GN} is only able to model pairwise interactive relations, hence it is consistency-unaware. Consequently \emph{GN} and \emph{modified GN} perform much worse than CAGNet. Albeit sharing the same feature extractor (Inception V3) with CAGNet, \emph{Joint + AS} and \emph{QP + CCCP} learn  human interactive relations via shallow structured models without incorporating higher-order contextual dependencies and consistency-aware reasoning, hence their performances are much worse than our CAGNet.

To provide a qualitative analysis of different models, we visualize a few predictions in Figure~\ref{fig:hiu_vis_result}. Though the predicted action labels are inconsistent or the predicted interactive relations violate the \emph{transitivity oracle} using either the \emph{Base-model} or the \emph{modified GN}, thanks to our proposed TOGN and CAR module, CAGNet is able to make perfect predictions, at least on these examples.

\section{Conclusion}
Under the observation that labeling consistencies across different atomic predictions are of great importance to achieve semantic and accurate understanding of human interactions, we have presented the so-called CAGNet which is able to resolve the labeling and grouping inconsistencies within HIU predictions. Our network relies on a TOGN module and a CAR module. The TOGN module addresses the inconsistency by learning better contextual features with higher-order graph networks,  while the proposed CAR module tackles the issue by exploiting the deductive reasoning bias of HIU explicitly. For efficient training and prediction,  we have cut the desired deductive reasoning into solving a surrogate energy-minimization, which reduces the chance of obtaining inconsistent HIU predictions and allows the training of all model parameters in an end-to-end way. Note that our CAR module is motivated by the HIU task, instead of  proposing a comprehensive system for deep logical reasoning.  Ablation study and comparison against the state-of-the-arts on three benchmarks have justified the effectiveness of the proposed approach.

{\small
\bibliographystyle{ieee_fullname}
\bibliography{hiu}
}

\end{document}